\documentclass{article}
\usepackage{arxiv}

\usepackage[ruled,linesnumbered]{algorithm2e}

\SetCommentSty{mycommfont}

\usepackage{graphicx}
\graphicspath{{Figures/}}
\usepackage{subfigure}

\usepackage{comment}
\usepackage{multirow}
\usepackage{bm}
\usepackage{enumitem}
\usepackage{soul}
\usepackage{float}
\usepackage{xcolor}
\usepackage{mathtools}
\usepackage{booktabs}

\usepackage{hyperref}

\newcommand{\wz}[1]{{\color{black}#1}}

\def\equationautorefname~#1\null{%
	Eqn.~(#1)\null
}

\newtheorem{definition}{Definition}

\title{NECA: Network-Embedded Deep Representation Learning
for Categorical Data}

\author{
 Xiaonan Gao \\
  University of Science and Technology Beijing\\
  \texttt{gaoxiaonan0001@163.com} \\
  \And
Sen Wu \\
  University of Science and Technology Beijing\\
  \texttt{wusen@manage.ustb.edu.cn} \\
  \And
  Wenjun Zhou \\
  University of Tennessee Knoxville\\
  \texttt{wzhou4@utk.edu} \\
}

\begin{document}
\maketitle

\begin{abstract}%
We propose NECA, a deep representation learning method for categorical data.
Built upon the foundations of network embedding and deep unsupervised representation learning, 
NECA deeply embeds the intrinsic relationship among attribute values and explicitly expresses data objects with numeric vector representations. 
Designed specifically for categorical data, NECA can support important downstream data mining tasks, such as clustering. 
Extensive experimental analysis demonstrated the effectiveness of NECA.

\end{abstract}

\keywords{unsupervised method, 
categorical attributes, network embedding}

\section{Introduction}

Categorical attribute datasets (CADs), defined as datasets where attributes are all categorical, are an important type of real-world data. 
CADs are collected at an increasing scale from many application domains, such as healthcare, human resources, and the Web. 
More broadly speaking, many existing numerical and mixed-type featured datasets can also be transformed into CADs. 
Therefore, general purpose methods are needed to provide a unifying framework of CADs to support their mining and analysis.

Categorical attributes have the known characteristics of values being enumerated, non-differentiable, and undefined algebraic operations. 
As a result, CADs have seen significant modeling challenges, since the proximity among data objects is hard to measure. 
Many of the well-known data mining methods rely on effectively measuring the proximity among instances. 
For data with solely numerical attributes, calculating proximity is straight-forward, for example, by using the Euclidean distance. 
However, measuring the proximity among CADs has often been done using ad-hoc, heuristic methods. 
A systematic extension of CAD representation and proximity measurement has been an understudied area. 

An example CAD in a talent recruitment context can be found in \autoref{tab:ex}, where each record represents a job applicant, and the attributes (``Gender,'' ``Specialty,'' and ``Position'') are all categorical.\footnote{``Name'' serves as an identifier variable (i.e., unique per record) in this toy example and is not used for modeling.} 
To cluster job applicants with categorical attributes, a commonly used technique is to code the proximity between two instances by each categorical variable into 1 if they are the same and 0 otherwise. 
Take the ``Position'' attribute as an example,
with domain knowledge, humans can understand that, compared with a ``Lawyer,'' a ``Programmer'' should have a closer proximity with a ``Technician.''
Exiting practice of coarse encoding missed the opportunity of 
discovering valuable information hidden in CADs, resulting in less accurate results. 
How to fully mine the actual relationship between the categorical attribute values, especially in an unsupervised way, 
has become a pressing need. 

\begin{table}[htp]
    \centering
    \caption{An example CAD used in talent analytics\label{tab:ex}}
    \begin{tabular}{cccc}
    \toprule
    Name & Gender & Specialty & Position \\
    \midrule
    John  & M & Engineering  & Programmer \\
    Tony  & M & Science      & Analyst \\
    Alisa & F & Liberal Arts & Lawyer \\
    Ben   & M & Engineering  & Programmer \\
    Abby  & F & Liberal Arts & Marketing \\
    James & M & Engineering  & Technician \\
    \bottomrule
    \end{tabular}
\end{table}


Existing related works on categorical data representation learning mainly include three categories: direct encoding-based methods, similarity-based methods, and embedding-based methods. 
Direct encoding-based methods only take into account the consistency and frequency relationship among categorical attribute values, ignore the latent semantic relationship. 
Similarity-based methods output the similarity matrix among the categorical data objects instead of an explicit representation of each data object. Thus, these methods can only be used for a subset of data mining techniques (i.e., those can use a similarity matrix as input). 
Embedding-based methods can overcome the limitations of the above two kinds to methods, but 
often relies on an external representation learning procedure, leading to 
susceptibility to the choice of those algorithms and their parameter configurations. 
Additionally, these embedding-based methods often require external information (e.g., labels). 


To develop a systematic way of CAD representation learning, this paper studies the unsupervised representation learning method of CAD, focusing on solving the aforementioned limitations in existing related works.
We leverage the network embedding to solve the problem of deep representation learning of CAD, and propose a network embedding based categorical data deep representation learning method called NEtwork-based CAtegorical representation learning (NECA). 
First, we convert the structured CAD in the form of a 2-dimensional table into a network form by constructing a weighted heterogeneous network. The attribute values are represented as the network nodes, and the relationships between the attribute values are expressed as edges, the strength of the relationship is reflected by the edge weight. Then, the relationship between the categorical attribute values is embedded into the deep numerical representation of node through network embedding. 
Finally, we can get the deep numerical representation of a categorical data object by concatenating the learned representations of attribute values describing the data object. In this way, the actual meaning relationship hidden in CADs can be reflected in the learned representation while the aforementioned limitations are not encountered.
Extensive experiments on UCI datasets show that our NECA can effectively learn the implicit correlation between different attribute values in the original CAD, and accurately represent it in the numerical representation, which conforms to the real data distribution and can obtain better performance of downstream mining task, like clustering.


The main contributions of this paper can be summarized as follows:
\begin{itemize}
    \item We introduce deep learning related theories into the research on unsupervised mining of structured CAD, and innovatively use network embedding to solve the categorical data representation learning problem. 

    \item A categorical network construction method is presented, which introduces all the useful information hidden in CADs into a weighted heterogeneous network.

    \item The intra-categorical attribute relationship and inter-categorical attribute relationship can be explored simultaneously through network embedding, which reveal the actual meaning hidden in real-world.

    \item NECA is a unsupervised learning model, which does not rely on any external information and will not be affected by other modelling processes.
\end{itemize}


The remainder of this paper is organized as follows. 
\autoref{sec:prob} provides the problem formulation and an overview of NECA. 
Subsequently, details of the proposed NECA framework are discussed in two separate sections. 
In \autoref{sec:network}, we present details about the heterogeneous network construction among intra- and inter-categorical attribute relationships. 
In \autoref{sec:learn}, we present the deep representation learning steps, which fuse the information learned from the previous step. 
\autoref{sec:algo} summarizes the NECA algorithm. 
We present experimental results in \autoref{sec:experiment} and discuss related work in \autoref{sec:relate}. 
Finally, we draw conclusions in \autoref{sec:conclusion}.

\section{Problem Definition and Overview\label{sec:prob}}

This section first provides preliminary information and the problem definition, and then offers an overview of our proposed solution. 

\subsection{Preliminaries and Problem Definition}

We first clarify a few basic concepts. 
A \emph{categorical} attribute is one that can take values from a finite, discrete set known as its \emph{domain}, 
where each value represents a category instead of a quantity. For instance, the domain of ``Gender'' could be written as $dom(Gender) = \{M, F, other\}$. 
The attribute ``Gender'' can only take 1 of the 3 possible values in its domain. 
Numerical summaries such as means and percentiles are not applicable. 
In this case, we can say that ``Gender'' is a categorical attribute.

Note that although categorical attributes can be further divided into nominal and ordinal attributes, 
in this paper, we limit our scope to nominal variables only. 
An ordinal variable will be treated the same way as nominal variables by ignoring 
the ordering information among attribute values. Similarly, numerical attributes may be preprocessed into categorical versions.
This makes our method more widely applicable as it does not require knowing the ordering information among attribute values.
Formally, we have the following definition of a categorical attribute dataset.

\begin{definition}[CAD\label{def:CAD}]
A categorical attribute dataset (CAD) is a set of $n$ data objects $X = \{\bm{x}_1,\bm{x}_2,\dots,\bm{x}_n\}$, 
where each data object is measured by a set of $m$ categorical attributes $\mathcal{C} = \{ C_1, C_2, \dots, C_m \}$.  
We can write $\bm{x}_i = (x_{i1},x_{i2},\dots,x_{im})$ as the $i$-th data object, where $x_{ij} \in dom(C_j)$, $\forall i = 1, 2, \dots, n$, $\forall j = 1, 2, \dots, m$.
\end{definition}

Given a CAD, without a pre-defined distance or similarity measure, it is impossible to quantify the proximity among categorical data objects. 
Widely used data mining techniques, such as K-means clustering and principal component analysis, therefore, cannot be readily applied to CADs. 
In this study, we propose a systematic way to learn numerical representations of categorical data objects, 
such that the intrinsic relationships among the data objects are well captured to support downstream data mining. 

With the goal of supporting downstream data mining tasks, the technical challenge of this representation learning problem is to output the CAD's numerical representation from the original categorical form. 
In essence, the task is to mine the implicit relationship among data objects in the CAD, to express all the useful information into the learned numerical representation.

\subsection{NECA: An Overview}

\begin{figure}[htp]
    \centering
    \includegraphics[width=0.7\linewidth]{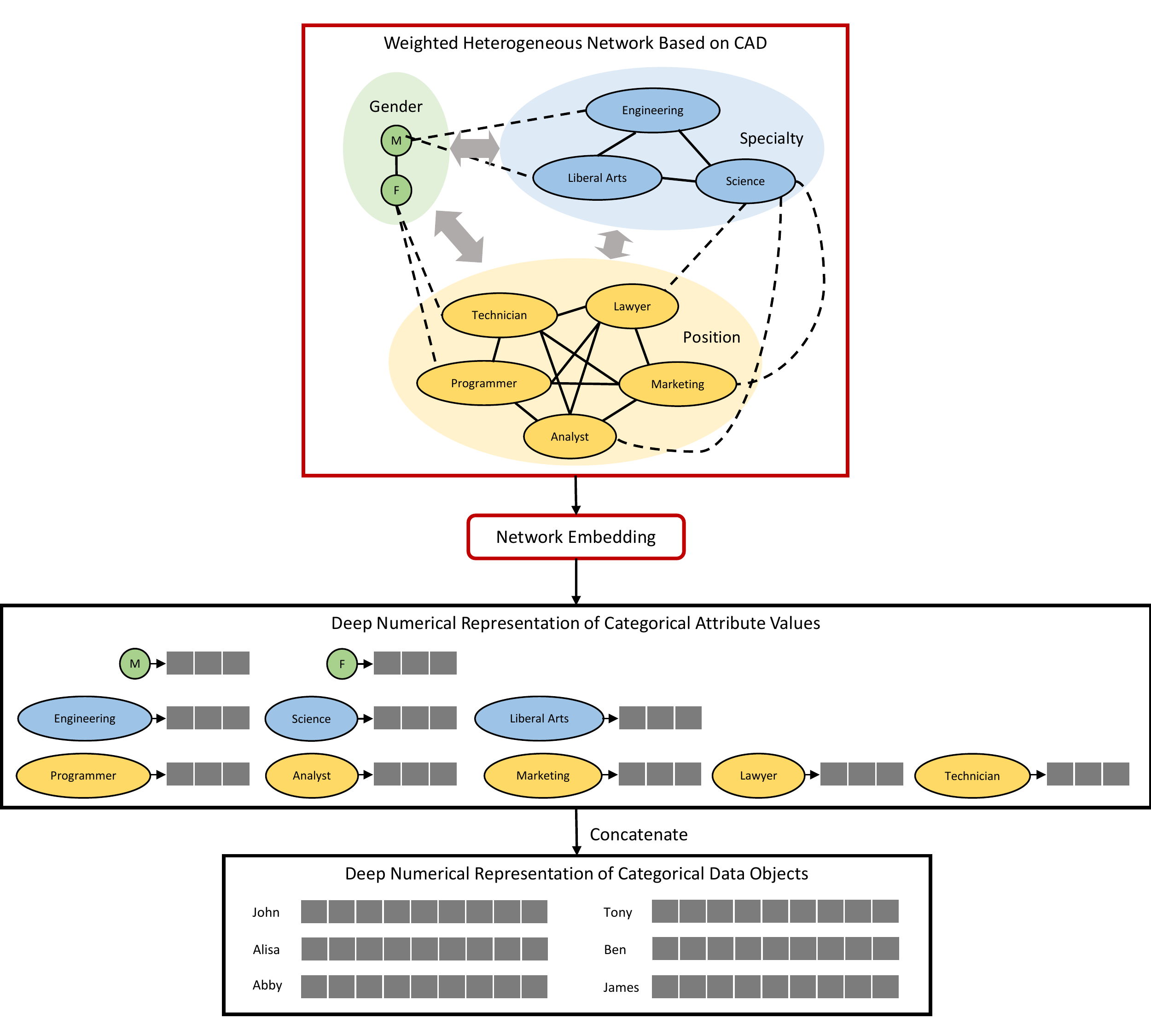}
    \caption{The basic idea of NECA.}
    \label{fig:3-3}
\end{figure}

To learn the effective numerical representations for CADs, in this paper, we propose a deep numerical representation learning model for categorical data based on network embedding, called NECA.
The basic idea of NECA is illustrated in \autoref{fig:3-3} with a stylized example, which consists of two phases.
In the first phase, we construct a weighted heterogeneous network to represent the two kinds of relationships among the categorical attribute values in CAD, one is the relationship of values between different attributes, the other is the relationship within the same attribute. The attribute values are treated as nodes, and their relationships are expressed in the weighted edges. Details of this phase can be found in \autoref{sec:network}.
In the second phase, we learn the deep numerical representation of each attribute value based on the constructed network through network embedding. In this way, the learned representation contains the potential useful information hidden in the original CAD. Along this line, we can get the deep numerical representations of categorical data objects by concatenating the learned representations of attribute values. More details of this phase are discussed in \autoref{sec:learn}.

\section{Network Construction\label{sec:network}}

To derive the deep numerical representations of CAD that capture the implicit relationship between attribute values in the original feature space, 
we first construct a weighted heterogeneous network using the attribute values as the nodes and their relationships as the weighted edges.

\begin{definition}[CAV Node]
Suppose that categorical attribute $C_j$ has domain $dom(C_j) = \{ c_j^1, c_j^2, \dots, c_j^{|c_j|}\}$, $1\le j\le m$. 
In other words, the categorical attribute $C_j$ has $|c_j|$ unique values. 
When attribute $C_j$ taking the $k$-th unique value, $c_j^k$, we say that 
$C_j=c_j^k$ makes a categorical attribute value (CAV) node, $1\le k\le |c_j|$.
We write the node as $c_j^k$ for short.
\end{definition}

\begin{definition}[CAV Node Set\label{def:CAVset}]
For a given CAD $X$, 
the CAV node set consists of all possible CAV nodes. 
Namely, 
\begin{equation}
V = \{ c_j^k | j = 1,2, \dots, m; k = 1, 2, \dots, |c_j| \}.    
\end{equation}
The total number of CAV nodes is $|V| = \sum_j |c_j|$. 
\end{definition}

\wz{Our goal is to construct the edges and formulate their weights to express relationship among the CAVs, which in turn can support developing the embedding among CAD objects.}

\begin{figure}[htp]
    \centering
    \includegraphics[width=0.6\linewidth]{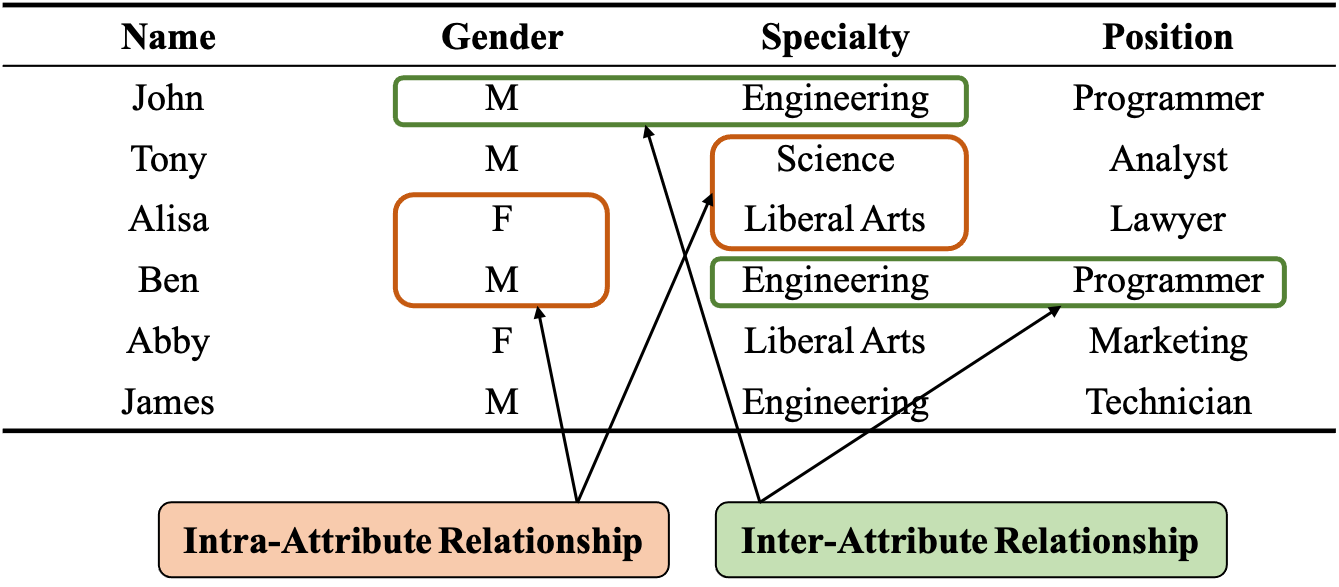}
    \caption{Example CAD to demonstrate attribute-value relationship between and within attributes.}
    \label{fig:3-2}
\end{figure}

We consider two types of relationships: IntEr-Categorical Attribute Relationship (IECAR) and IntrA-Categorical Attribute Relationship (IACAR). 
An intuitive example is shown in \autoref{fig:3-2}.
The two green boxes in \autoref{fig:3-2} demonstrate inter-attribute relationship between attribute-value pairs.
In this example dataset, the gender of all instances whose specialty is Engineering are Male, indicating that the relationship between ``Engineering (Specialty)'' and ``M (Gender)'' is stronger than between ``Engineering (Specialty)'' and ``F (Gender).'' 
In addition, ``Engineering (Specialty)'' and ``Programmer (Position)'' frequently co-occur, indicating that these two CAVs are closely related. 
For the intra-attribute relationship within the same attribute, as labeled in red boxes in \autoref{fig:3-2} for illustration, examines the relationship between different values within the same attribute. 
The frequency of ''M (Gender)'' is higher than that of ''F (Gender)'', indicating that the when ''F (Gender)'' appears, more rare information will be provided. 
Similarly, in the relationship ''Science (Specialty) - Liberal Arts (Specialty)'', there is more information in ''Science (Specialty)'' than ''Liberal Arts (Specialty)''.

We will construct the weighted heterogeneous network based on the IECAR and IACAR relationships respectively.

\begin{figure*}
    \centering
    \subfigure[Gender vs. Specialty]{\includegraphics[width=0.3\linewidth]{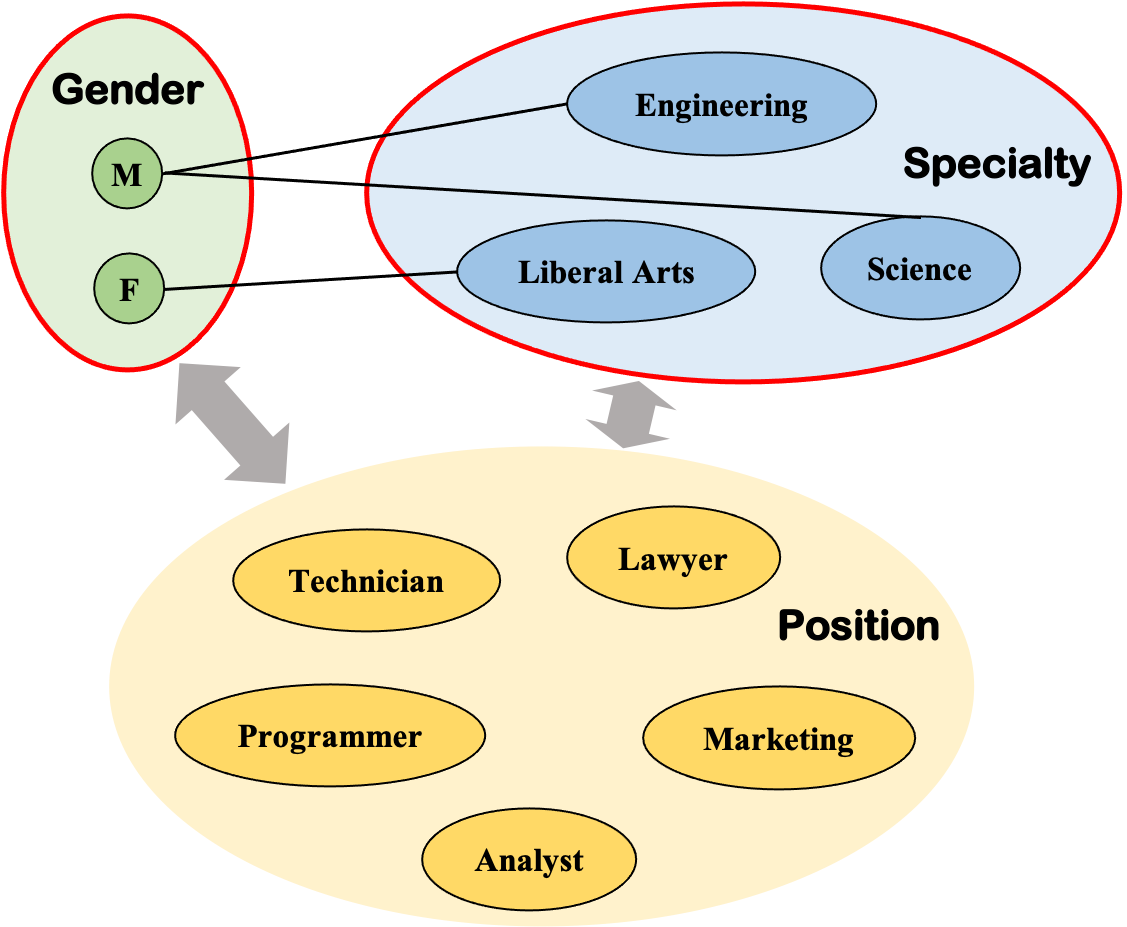}}
    \hfill
    \subfigure[Gender vs. Position]{\includegraphics[width=0.3\linewidth]{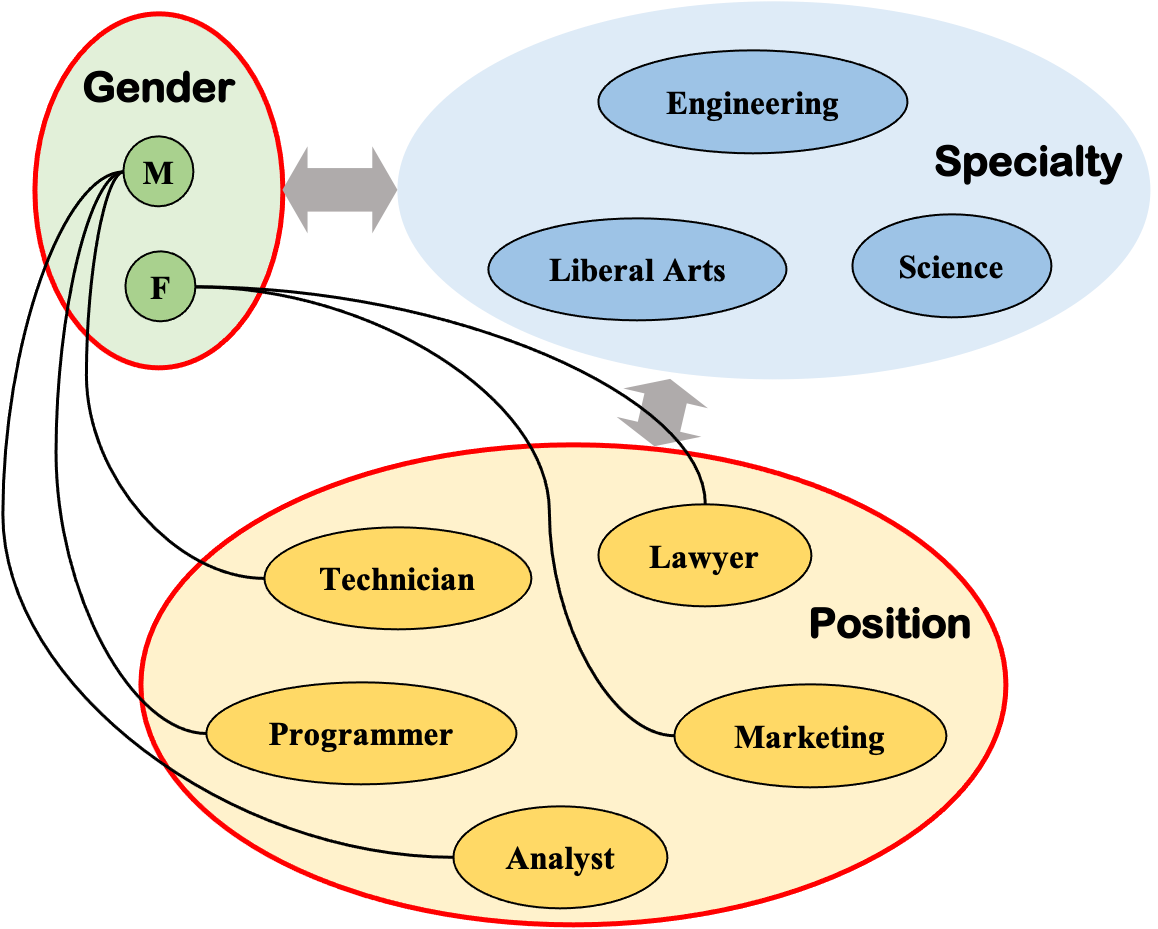}}
    \hfill
    \subfigure[Specialty vs. Position]{\includegraphics[width=0.3\linewidth]{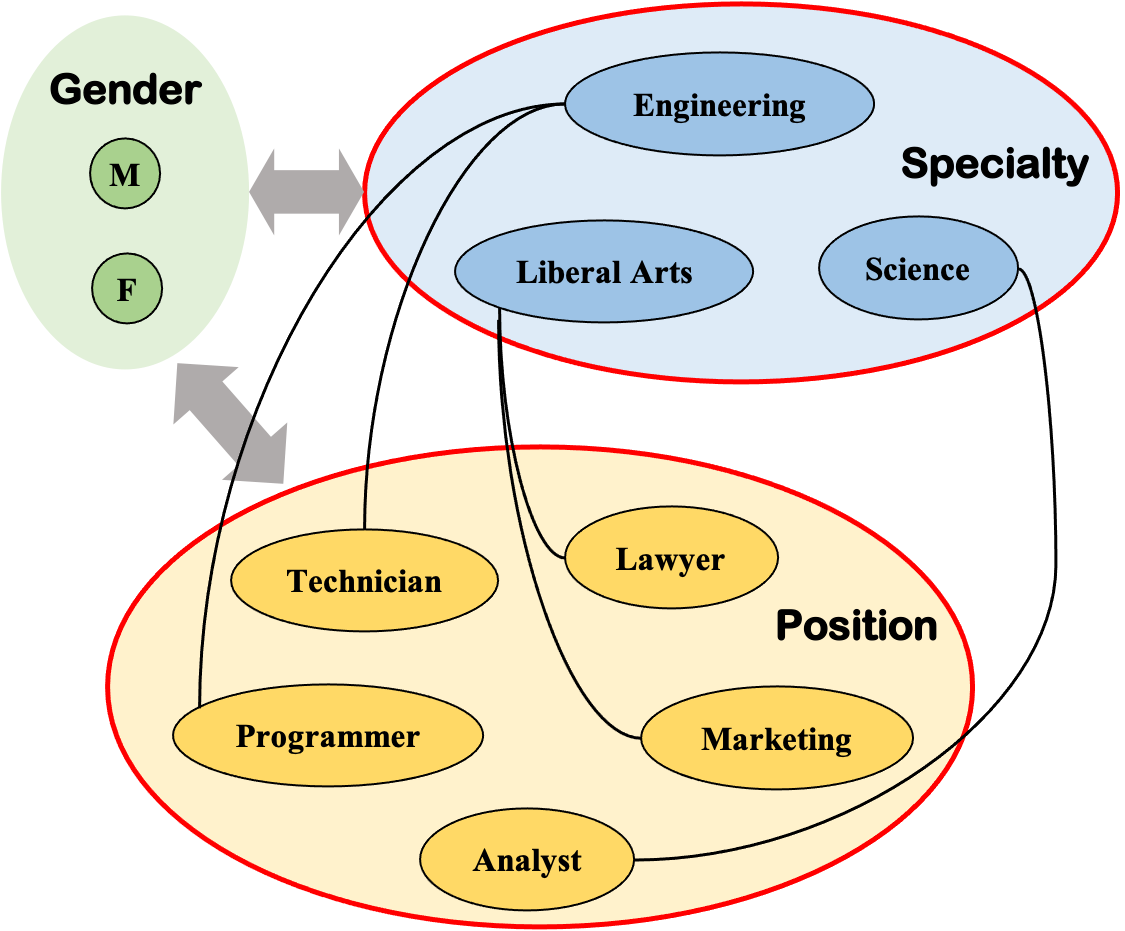}}
    \caption{Network diagrams illustrating IECAR.\label{fig:3-4}}
\end{figure*}

\subsection{IECAR Network Construction}

Based on the CAD example in \autoref{fig:3-2}, \autoref{fig:3-4} shows a schematic diagram of the network constructed based on the IECAR relationship. If two attribute values have a co-occurrence relationship in the dataset, they are connected by an edge, otherwise the two attribute values are boundlessly connected. The weight on an edge can be calculated by \autoref{eqn:3-3} and \autoref{eqn:3-4}. 

\begin{definition}[IECAR Weights\label{def:IECARweights}]
The IECAR weight on edge $e(c_j^l, c_{j'}^{l'})$ is calculated as
\begin{equation}\label{eqn:3-3}
    w_{inter} (c_j^l, c_{j'}^{l'}) = \frac{ \exp \{ f(c_j^l, c_{j'}^{l'}) \} }{ \sum_{ e(c_j^l, c_{j'}^{l'}) \in E_{inter} } \exp \{ f(c_j^l, c_{j'}^{l'}) \} },
\end{equation}
where 
\begin{equation}\label{eqn:3-4}
    f(c_j^l, c_{j'}^{l'}) = \left| \{ \bm{x}_i | x_{ij}=c_j^l \land x_{ij'}=c_{j'}^{l'} \} \right|
\end{equation}
represents the number of co-occurrences of attribute values $c_j^l$ and $c_{j'}^{l'}$ in the dataset $X$. 
\end{definition}

The IECAR weights $w_{inter} (c_j^l, c_{j'}^{l'})$ represents the weight of the edge between $c_j^l$ and $c_{j'}^{l'}$, which is normalized with the softmax function \cite{Ref43}.

\begin{definition}[IECAR Network]
Given a CAD $X$, the weighted network 
constructed using IECAR can be expressed as
\begin{equation}
    G_{inter}=(V,E_{inter}),
\end{equation}
where $V$ is the CAV node set of $X$ (see \autoref{def:CAVset})
and $E_{inter}$ are the set of edges, whose weights are specified in \autoref{def:IECARweights}. 
\end{definition}

This weight calculation method can reflect the relationship between values belonging to different attributes in the network. The greater the number of co-occurrences, the closer the relationship between attribute values should be, and the greater the weight of the corresponding edge.

\subsection{IACAR Network Construction}

This subsection describes how we operationalize the network construction based on IACAR.

\begin{figure}[htp]
    \centering
    \includegraphics[width=0.56\linewidth]{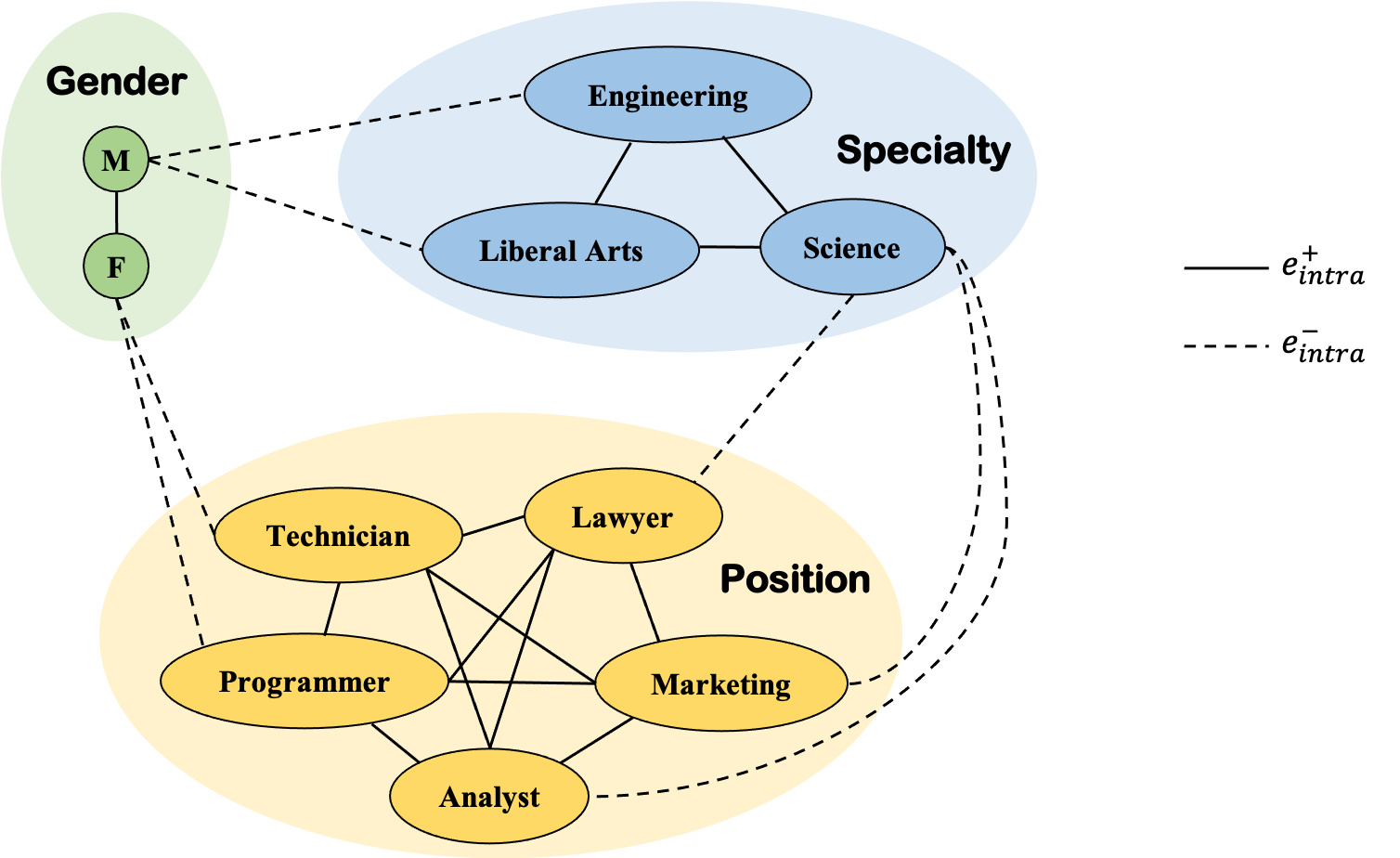}
    \caption{Network diagram illustrating IACAR.}
    \label{fig:3-5}
\end{figure}

According to the CAD example in \autoref{fig:3-2}, \autoref{fig:3-5} shows a schematic diagram of the network based on the relationship within the categorical attribute. The value nodes in the attributes ``Gender``, ``Specialty'' and ``Position'' are connected in pairs, represented by solid lines. In addition, to ensure the connectivity of the network, each node is randomly connected to a value node in other attribute, which is represented by a dotted line. In order not to affect the accuracy of the relationship mining within the categorical attributes, these randomly added edges are given smaller weights.

\begin{definition}[IACAR Weights\label{def:IACAR}]
The IACAR weight on edge $e(c_j^l, c_{j'}^{l'})$ is calculated as
\begin{equation}\label{eqn:3-7}
    w_{intra} (c_j^l, c_{j'}^{l'}) = \frac{ \exp \{ h(c_j^l, c_{j'}^{l'}) \} }{ \sum_{ e(c_j^l, c_{j'}^{l'}) \in E_{intra} } \exp \{ h(c_j^l, c_{j'}^{l'}) \} }
\end{equation}
where 
\begin{equation}\label{eqn:3-8}
h(c_j^l, c_{j'}^{l'}) = \left\{
\begin{array}{ll}
    \frac{ n }{g(c_j^l) + g(c_j^{l'})}, &  \textrm{if } j=j'; \\
    \beta, & \textrm{otherwise}.
\end{array}
\right. 
\end{equation}
where $n$ is the number of instances in the CAD $X$, 
$g(c_j^l) = | \{ \bm{x}_i | x_{ij}=c_j^l, 1\le i \le n\}|$ represents the number of instances in $X$ whose attribute $C_j$ takes the value $c_j^l$, 
and $\beta$ is a small positive coefficient to ensure connectivity among the CAV nodes. 
\end{definition}

\begin{definition}[IACAR Network]
Given a CAD $X$, the weighted network constructed using IACAR can be expressed as
\begin{equation}
    G_{intra}=(V,E_{intra}),
\end{equation}
where $V$ is the CAV node set of $X$ 
and $E_{intra}$ 
consists of the edges among the CAVs, whose IACAR weights are specified in \autoref{def:IACAR}. 
\end{definition}

\section{Representation Learning\label{sec:learn}}

Based on the weighted heterogeneous networks constructed for the CAD, as described in the previous section, we learn the deep numerical representations that best capture the inter-attribute and intra-attribute relationships in the CAD. 

\subsection{Within-Network Relationship Learning}
\label{subsec-within network learn}

The IACAR and IECAR learning steps follow a similar process. The main difference between the two learning tasks is that they are learned within the IACAR and IECAR networks, respectively. 

Take IECAR learning as an example, for a given CAV $c_j^l$, mining its relationship with other values in the different categorical attributes is based on $G_{inter}=(V,E_{inter})$. 
The basic idea is to explore the influence of neighbor nodes connected to the target node $c_j^l$ through $E_{inter}$, and quantify this influence as the neighbor node’s contribution to the target node. 
The general process is as follows. 
First, we identify the neighbor node set for each CAV node. 
Then, based on the relationships between each node (called the target node) and its neighbor nodes, we use the attention mechanism \cite{Ref151} to learn the contribution of the neighbor nodes on the target node. In this way, we can obtain the representation of the target node within the IECAR network, which can reflect the IECAR relationship. Learning the representation based on IACAR network is the same.

\begin{definition}[Neighbor Node Set]
In a network $G$, for any given node $v$, its neighbor nodes make a neighbor node set, represented as $\Omega_G(v) = \{v' | weight (v, v') > 0 \}$.
\end{definition}

Given the above definition, while learning IACAR relationship, the neighbor node set on the IACAR network is $\Omega_{intra}(v)$. Similarly, the neighbor node set on the IECAR network is $\Omega_{inter}(v)$. 

Take IACAR network for example, the relationship between the target node $c_j^l$ and its neighbor nodes $\Omega_{intra}(c_j^l)$ is explored using the attention mechanism. 
First, we map the nodes into the same feature space:
\begin{equation}\label{eqn:3-9}
h_j^l = W_1 \cdot c_j^l,
\end{equation}
where $W_1$ is used to map the node representing attribute value into the latent feature space, and 
$h_j^l$ is the node $c_j^l$'s representation in that space.
After projecting both the target node $c_j^l$ and its neighbor nodes $c_j^{l'} \in \Omega_{intra}(c_j^l)$ into the same feature space, we can calculate the contribution of $c_j^{l'}$ to $c_j^{l}$ as:
\begin{equation}\label{eqn:3-10}
\delta(c_j^{l}, c_j^{l'}) = LeakyReLU(r_{intra} \cdot [h_j^{l} || h_j^{l'}]),
\end{equation}
where $r_{intra}$ and $W_1$ are parameters to learn, the $||$ operator concatenates the projected representations, $LeakyReLU$ \cite{Lakshmi2021ALB} is the activation function, which is used to introduce nonlinear factors and enhance the learning ability of the network.

Using \autoref{eqn:3-10} we can compute the contribution of all neighbor nodes in $\Omega_{intra}(c_j^l)$ to target node $c_j^{l}$. Then, we use the softmax function \cite{Ref43} to normalize them and obtain the weight as \autoref{eqn:3-11}.

\begin{definition}[Neighbor Node Weight]
Suppose that the neighbor nodes of a target node $c_j^l$ is $c_j^{l'} \in \Omega(c_j^l)$. 
The impact of the neighbor node $c_j^{l'}$ on target node $c_j^l$ can be calculated as 
\begin{equation}\label{eqn:3-11}
\alpha(c_j^{l}, c_j^{l'}) = \frac{ exp\{\delta(c_j^{l}, c_j^{l'})\} }{ \sum_{c_j^{l'} \in \Omega(c_j^l)} exp\{\delta(c_j^{l}, c_j^{l'})\} },
\end{equation}
where $\delta(c_j^{l}, c_j^{l'})$ can be found in \autoref{eqn:3-10}. 
\end{definition}

It is worth mentioning that the weights obtained using \autoref{eqn:3-10} and \autoref{eqn:3-11} are only dependent on the features of the target node and its neighbor nodes themselves.
Additionally, the mutual influence between any two nodes are asymmetric, which indicates that the weight from a neighbor node $c_j^{l'}$ to the target node $c_j^l$ is not always equal to the weight from the target node $c_j^l$ to the neighbor node $c_j^{l'}$.
The computing result depends on the ordering in the $||$ operator and the normalization. 

Still take IACAR network as an example, after obtaining the weights of all neighbor nodes $c_j^{l'}\in \Omega_{intra}(c_j^l)$ on the target node $c_j^l$, 
we can compute the target node's representation as follows:
\begin{equation}\label{eqn:3-12}
a_j^{l} = \sigma \left(
 \sum_{c_j^{l'} \in \Omega_{intra}(c_j^l)}
 \alpha(c_j^{l}, c_j^{l'}) \cdot h_j^{l'}
\right),
\end{equation}
where $\sigma$ is the activation function, which we specified as the ELU function \cite{Ref43}. 

\autoref{eqn:3-12} indicates that the learned representation based on the intra-attribute relationship of the target node can capture its neighbor nodes' inherent characteristics and their contribution on the target node. In other words, $a_j^l$ can fully reflect the relationship between the target node and its neighbor nodes.

Since \autoref{eqn:3-10}, \autoref{eqn:3-11}, and \autoref{eqn:3-12} are all based on \autoref{eqn:3-9} to characterize potential features of nodes, if the learned feature space is not suitable for the IACAR relationship exploration, all subsequent calculation processes will be affected. To solve this problem, the multi-head attention mechanism \cite{Ref151} is used to repeat the above operation $K$ times, which means that the IACAR relationship can be learned in $K$ different feature spaces, and $K$ numerical representations of each CAV is obtained in each space. Finally, the final representation of the target node is calculated through the concatenation operation based on the $K$ spaces:
\begin{equation}\label{eqn:3-13}
a_j^{l} = ||_{k=1}^K \{a(k)_j^{l} \},
\end{equation}
where $a(k)_j^{l}$ is the IACAR representation of target node $c_j^{l}$ learned in the $k$th space, which is calculated using \autoref{eqn:3-12}.

Similarly, from the IECAR network, we can obtain
\begin{equation}\label{eqn:3-13b}
e_j^{l} = ||_{k=1}^K \{e(k)_j^{l} \},
\end{equation}
where $e(k)_j^{l}$ is the IECAR representation of $c_j^{l}$ in the $k$th space based on the IECAR network.

Consequently, this subsection learns the deep numerical representations of nodes by mining the IACAR and IECAR relationships in CAD. Specifically, with attention mechanisms, in $K$ different feature spaces, the numerical representation of target node is learned considering the relationship from its neighbors respectively, and the final representation is obtained by concatenation operation.

\subsection{Between-Network Fusion}
\label{subsec-between network fusion}

After mining the IECAR and IACAR relationships, the two representations should be merged to learn the final numerical representation of CAV. The basic idea is to use the attention mechanism to learn the contribution of the IECAR and IACAR relationships on the CAV respectively, and then calculate the final representation of the CAV based on the quantified contribution weights.

First, we measure the importance of the IECAR and IACAR for the overall CAD. 
Based on the weighted heterogeneous network $G=(V,E)$, where $E=\{E_{inter},E_{intra}\}$, the attention mechanism is used to learn the importance of IECAR and IACAR for each CAV node.
Then, by calculating the average values, we obtain
the importance of IECAR and IACAR for the CAD $X$ as follows:
\begin{eqnarray}
\gamma_{inter} = \frac{1}{|V|} \underset{c_j^l \in V}{\sum} s \cdot \tanh(W_2 \cdot e_j^l + b),\label{eqn:3-17}\\
\gamma_{intra} = \frac{1}{|V|} \underset{c_j^l \in V}{\sum} s \cdot \tanh(W_2 \cdot a_j^l + b),\label{eqn:3-18}
\end{eqnarray}
where $\gamma_{inter}$ and $\gamma_{intra}$ are calculated based on all CAVs reflecting the importance of IECAR and IACAR relationship on the whole CAD dataset $X$, $s$, $W_2$, and $b$ are parameters to be learned. $W_2$ and $b$ will be used to map the IECAR and IACAR representations into the same feature space. $s$ is used to guide the learning of the importance score.

Subsequently, we normalize $\gamma_{inter}$ and $\gamma_{intra}$ as:
\begin{eqnarray}
\beta_{inter} = \frac{ \exp\{\gamma_{inter}\} }{ \exp\{\gamma_{inter}\} + \exp\{\gamma_{intra}\} },\label{eqn:3-19}\\
\beta_{intra} = \frac{ \exp\{\gamma_{intra}\} }{ \exp\{\gamma_{inter}\} + \exp\{\gamma_{intra}\} },\label{eqn:3-20}
\end{eqnarray}
where $\beta_{inter}$ and $\beta_{intra}$ respectively reflect the contribution weights of IECAR and IACAR on the CAD $X$. 

To this end, we can fuse the two types of relationships to obtain the final deep numerical representation of the target CAV $c_j^l$ as:
\begin{equation}\label{eqn:3-21}
f_j^l = \beta_{inter} \cdot e_j^l + \beta_{intra} \cdot a_j^l.
\end{equation}

Based on the numerical representations learned for CAVs, we can obtain the deep numerical representation of the CAD by concatenating the CAV representations 
\cite{Ref151}. 
\begin{equation}\label{eqn:3-27}
x'_i = \underset{ x_i^j=c_j^l, 1\le j\le m }{||} f_j^l.
\end{equation}


\subsection{Learning Loss}
\label{subsec-learning loss}

For training the network, we hold the assumption that nodes with similar neighbors are similar to each other.
Take a CAV node $c_j^l$ and its neighbor node $c_{j'}^{l'} \in \Omega_{inter} (c_j^l)$ for example, we first define the impacting strength of $c_{j'}^{l'}$ on $c_j^l$ as :
\begin{equation}\label{eqn:3-25}
p(c_{j'}^{l'}|c_j^l) = \frac{ w_{inter} (c_j^l, c_{j'}^{l'}) }{ \underset{c_{j'}^{l''} \in \Omega_{inter}(c_j^l)}{\sum} w_{inter} (c_j^l, c_{j'}^{l''}) },
\end{equation}
where the numerator $w_{inter}(c_j^l, c_{j'}^{l'})$ is the weight of the edge connecting between the target node $c_j^l$ and its neighbor node $c_{j'}^{l'}$ in the IECAR network, and the denominator is the sum of the edge weights between all neighbor nodes connecting to the target node. 
To preserve the proximity, the similarity of the learned numerical representations for $c_j^l$ and $c_{j'}^{l'}$ should be close to $p(c_{j'}^{l'}|c_j^l)$.

\begin{definition}[NECA Learning Loss\label{def:LossInter2}]
Given the learned numerical representations of all CAVs, the NECA learning loss can be computed as
\begin{equation}\label{eqn:3-26}
Loss(W) = -\frac{1}{|E_{inter}|} \underset{e(c_j^l, c_{j'}^{l'}) \in E_{inter}}{\sum}{\tilde{l}(c_j^j, c_{j'}^{l'})},
\end{equation}
where
\begin{eqnarray*}
\tilde{l}(c_j^j, c_{j'}^{l'}) &= & p (c_{j'}^{j'}| c_j^l) \log G(f_j^l, f_{j'}^{l'}) + \\
&& (1- p (c_{j'}^{j'}| c_j^l)) \log (1- G(f_j^l, f_{j'}^{l'}) )
\end{eqnarray*}
and $G(f_j^l, f_{j'}^{l'})$ refers to the Gaussian kernel similarity between $f_j^l$ and $f_{j'}^{l'}$. 
\end{definition} 
By minimizing the loss and updating parameters using Adam \cite{Kingma2015AdamAM}, we can learn the deep numerical representations of CAVs, which contain the potential valuable information hidden in CAD.

\section{The NECA Algorithm\label{sec:algo}}

NECA transforms the original CAD into a weighted heterogeneous network, and uses the idea of network embedding to fully mine the relationship between the CAVs in the dataset, including IECAR and IACAR. 
The learned representations of CAVs are concatenated to produce the deep numerical representations of the CAD. 
This section summarizes the detailed steps of our NECA algorithm.

\subsection{Algorithm Description}

The specific steps of NECA is summarized in \autoref{alg:3-1}. 

\begin{algorithm}[h]
\caption{NEtwork-based CAtegorical representation learning (NECA)\label{alg:3-1}}
\SetAlgoLined
\SetKwInOut{Input}{Input}
\SetKwInOut{Output}{Output}
\SetKwInOut{Param}{Param}

\Input{$X$: An input CAD as defined in \autoref{def:CAD}.}
\Output{$X'=\{\bm{x}'_1,\bm{x}'_2,\dots,\bm{x}'_n\}$: numerical representations}

\Param{$K$: The number of heads in the multi-head attention mechanism (default: 8);
}

\tcp{Initialize the IECAR and IACAR networks}
$G \leftarrow ConstrucHetNet(X)$\;

\tcp{Learn numerical representations of CAVs}
\While{the training stop condition is not satisfied}{
    \tcp{Within-network relationship learning}
    $A' \leftarrow DeepLearnCAV(G_{intra})$\;
    $E' \leftarrow DeepLearnCAV(G_{inter})$\;
    
    \tcp{Between-network fusion}
    Calculate the contribution weights of IECAR and IACAR relationship on $X$: $\beta_{inter}$, $\beta_{intra}$\;
    Fuse $A'$ and $E'$ to obtain the deep numerical representation $F=\{f_1^1, f_1^2, \dots, f_m^{|c_m|}\}$ of all CAVs\;
    Calculate the loss function: $Loss$\;
    Update the parameters in NECA;
}

\tcp{Learn numerical representations of CAD}
Obtain the deep numerical representation $\bm{x}'_i$ of data object $\bm{x}_i$, by concatenating the learned representations of CAVs

\Return{$X'$}
\end{algorithm}

\begin{procedure}[h]
\caption{ConstrucHetNet(X)}
\tcp{Create the node set of network}
\label{proce-consNet}
$V \leftarrow CAVs$\;
\tcp{Create the IECAR network}
\For{$c_j^l$ and $c_{j'}^{l'} \in V$, $j \neq j'$}{
    \If{there is $x_i \in X$, $x_i^j=c_j^l$ and $x_i^{j'}=c_{j'}^{l'}$}{
        Add edge $e_{inter}(c_j^l, c_{j'}^{l'})$\;
        Calculate edge weight $w_{inter}(c_j^l, c_{j'}^{l'})$\;}
}
Normalize the weights in IECAR network\;
\tcp{Create the IACAR network}
\For{$c_j^l$ and $c_{j}^{l'} \in V$, $l \neq l'$}{
        Add edge $e_{intra}(c_j^l, c_j^{l'})$\;
        Calculate edge weight $w_{intra}(c_j^l, c_j^{l'})$\;
}
\For{$c_j^l \in V$}{
    Randomly select a node from other attribute: $c_{j'}^{l'}, j'\neq j$\;
    Add edge $e_{intra}(c_j^l, c_{j'}^{l'})$ and calculate $w_{intra}(c_j^l, c_{j'}^{l'})$\;
}
Normalize the weights in IACAR network\;
\Return{$G=\{V,E\}, E=\{E_{inter}, E_{intra}\}$}
\end{procedure}

\begin{procedure}[h]
\caption{DeepLearnCAV(G)}
Initialize the numerical representations of all nodes $c_j^l$, $1 \le j \le m$, $1 \le l \le |c_j|$, through One-hot encoding\;
\tcp{Construct the $K$ heads}
\For{$c_j^l \in V$}{
    \For{$k\in \{1,\dots,K\}$}{
        Find out the neighbor node set of $c_j^l$ in the $G$: $\Omega(c_j^l) \in \{ \Omega_{inter}(c_j^l), \Omega_{intra}(c_j^l) \}$\;   
        \For{$c_j^{l'}\in \Omega(c_j^l)$}{
             Calculate the contribution of $c_j^{l'}$ to $c_j^l$: $\alpha(c_j^l,c_j^{l'})$\;
        }
        Calculate the numerical representation $v(k)_j^l \in \{a(k)_j^l,e(k)_j^l\}$ of $c_j^l$ in this head\;
    }
    Obtain the numerical representation of $c_j^l$: $v_j^l=||_{k=1}^K \{v(k)_j^{l} \}$\;
}
$V' = \{v_j^l, 1\le j\le m, 1\le l\le |c_j| \}$\;
\Return{$V'$}
\end{procedure}


NECA first constructs a weighted heterogeneous network based on the IECAR and IACAR relationships,
where CAVs make the nodes, and the relationships between different CAV nodes make the edges, whose weights are to be formulated (Line 1). 

Then, using the idea of network embedding, we learn the deep numerical representation of each node in the network (Lines 2 to 9). This representation can reflect the topological relationship in the network, and mine the relationship between the categorical attributes and the relationship within the attributes, which means that the learned representation of a CAV contains all the relations related to other CAVs. By concatenating the learned representations of CAVs contained in a certain data object, the deep numerical representation of the data object can be obtained (Line 10). Specifically, the training stop condition can be that the epoch is larger than a specified value, or the loss is convergent.

\subsection{Characteristics of NECA}

The NECA algorithm has the following characteristics.

(1) The deep numerical representation learning process of CAD is only based on the dataset, and does not need to use external information such as class labels. It belongs to unsupervised mining and is suitable for the common unlabeled CAD in real-world.

(2) The IECAR and IACAR relationships between CAVs can be fully mined and expressed in the learned deep numerical representation, based on the idea of network embedding.

(3 The learned representation is independent to other learning procedures and not impacted by the parameter settings of other learning tasks.

(4) To the best of our knowledge, this is the first time to learn the numerical representations of CAD using the basic idea of network embedding, which provides a new way to solve the representation learning problem of CAD.

\section{Experimental Results\label{sec:experiment}}

In this section, we first introduce the experimental setup and then present and discuss the results. 

\subsection{Experimental Setup}

This subsection provides the basic information of our experimental setup, including datasets, experimental design, parameter setting, platform and implementation. 

\textbf{Datasets}. We use 11 publicly available CADs from the UCI Machine Learning Repository \cite{Dua:2019} for experimental analysis. 
Summary statistics of each dataset are shown in \autoref{tab:3-2}. 
The datasets cover a wide range of sample sizes (\#Cases), dimensionality (\#Attr.), and number of subgroups (\#Classes). 
In their original version, there were missing values in the BC, DE, MA, MU, and PT datasets, 
which were imputed using attribute-specific modes.

\begin{table}[htp]
\centering
\caption{Experimental Datasets\label{tab:3-2}}
\begin{tabular}{lrrr}
\toprule
Name & \#Cases & \#Attr. & \#Classes \\
\midrule
Breast Cancer (BC) & 286 & 9 & 2 \\
Car Evaluation (CE) & 1,728 & 6 & 4 \\
Dermatology (DE) & 366 & 35 & 6 \\
Lymphography (LY) & 148 & 18 & 4 \\
Mammographic (MA) & 961 & 5 & 2 \\
Mushroom (MU) & 5,644 & 22 & 2 \\
Primary Tumor (PT) & 339 & 17 & 17 \\
Soybean Small (SB) & 47 & 35 & 4 \\
Spect Heart Train (SH) & 80 & 22 & 2 \\
Wisconsin (WI) & 699 & 9 & 2 \\
Zoo (ZO) & 101 & 16 & 7 \\
\bottomrule
    \end{tabular}
\end{table}

\textbf{Benchmark Methods}.
Our benchmark methods include representation learning methods and network embedding methods. 
The representation learning methods are 
\begin{itemize}
    \item Frequency-based (FQ) encoding \cite{Ref35};  
    \item One-hot (OH) encoding \cite{Ref33}; 
    \item Coupled data embedding (CDE) \cite{Ref40}. 
\end{itemize}
In specific, FQ and OH are the commonly used categorical embedding models. CDE is the newly proposed categorical data representation learning algorithm in recent years, whose superiority over other models has been proved. Therefore, we believe that the experiments based on these baselines can validate the effectiveness of NECA.

The network embedding benchmarks are based on DeepWalk \cite{Perozzi2014DeepWalkOL} and Node2vec \cite{Grover2016node2vecSF}. 
Since these methods are designed for homogeneous networks, we adapt them for the IACAR and IECAR networds, respectively.
\begin{itemize}
    \item The DeepWalk embedding based on IACAR (DWa);
    \item The DeepWalk embedding based on IECAR (DWe); 
    \item The Node2vec embedding based on IACAR (NVa);
    \item The Node2vec embedding based on IECAR (NVe).
\end{itemize}

\textbf{Parameter Setting}. The number of heads in the multi-attention mechanism is set as 8. 
Regarding the benchmark methods, 
there is no need to set parameters in OH and FQ. 
The CDE needs to set parameter $\alpha$, which is used to determine the number of clusters, and parameter $\beta$, which is used for dimension reduction. In our experiments, we set $\alpha=10$, $\beta=10^{-10}$ following the default parameters in the original publication \cite{Ref40}. 

\textbf{Platform and Implementation}. The experiments are conducted on a Windows Server 64-bit system (4-CPU, each with 2.5GHz with Quad-Core, and 16G main memory). All algorithms are implemented in Python.

\begin{table}
\centering
\caption{Internal Validation Results by CH\label{tab:3-3}}
\begin{tabular}{l|rrr|rrrr|r}
\toprule
& FQ & OH & CDE & DWa & DWe & NVa & NVe &  NECA \\ 
\midrule
BC & 7.59 & 6.95 &                \underline{\it 9.63}  & \textbf{ 10.3} & 5.12 &                6.72  &                8.74    & 9.16 \\ 
CE & 30.3 & 32.0 &                18.7  & \underline{\it 39.9} & 10.9 &                17.9  &                26.6    & \textbf{46.8} \\ 
DE & 30.5 & 33.0 & \textbf{ 77.23} &                35.0  & \underline{\it 73.5} &                31.8  &                73.1    & 45.2 \\ 
LY & 6.31 & 5.98 &                7.41  & \underline{\it 7.84} & 5.18 &                5.58  &                6.05    & \textbf{8.66} \\ 
MA & 81.4 & 126. &        \textbf{247.} &                103.  & 105. & \underline{\it 204.} &                100.    & 44.9 \\ 
MU & 761. & 885. & \underline{\it 1269.} &                982.  & 675. &                1040.  &                978.    & \textbf{1396.} \\ 
PT & 4.99 & 5.03 & \textbf{ 5.94} &                4.79  & 4.33 &                4.78  &                \underline{\it 5.63}    & 5.03 \\ 
SB & 14.5 & 16.1 & \underline{\it 36.4} &                14.5  & 17.0 &                21.4  &                20.2    & \textbf{60.4} \\ 
SH & 6.91 & 6.98 & \underline{\it 6.99} &                6.76  & 6.72 &                5.76  &                6.38    & \textbf{7.53} \\ 
WI & 217. & 164. & \textbf{ 460.} &                160.  & 210. &                160.  &                188.    & \underline{\it 263}. \\ 
ZO & 31.1 & 31.5 &                36.8  &                28.0  & 33.1 &                30.6  & \underline{\it 40.8}   & \textbf{55.6} \\ 
\bottomrule
\end{tabular}
\end{table}

\begin{table}
    \centering
    \caption{Internal Validation Results by S\label{tab:3-4}}
\begin{tabular}{l|rrr|rrrr|r}
\toprule
& FQ & OH & CDE & DWa & DWe & NVa & NVe & NECA \\ 
\midrule
BC &  0.04 &                 0.04  &                 \textbf{ 0.06}  &  \textbf{ 0.06} &          0.04  &                 0.04  &                 \textbf{ 0.06}    &          \underline{\it 0.05} \\
CE & -0.01 &                -0.00  &                -0.09  &                -0.05  &  \textbf{0.01} &                -0.10  &  \underline{\it 0.00}  & -0.01 \\
DE &  0.08 &                 0.10  &  \textbf{ 0.19} &                 0.11  &          \textbf{0.19}  &                 0.09  &                 \underline{\it 0.18}   &  0.09 \\
LY &  \underline{\it 0.06} &                 \underline{\it 0.06}  &  \underline{\it 0.06} &                 \underline{\it 0.06}  &          0.03  &                 0.05  &                 0.03   &  \textbf{0.07} \\
MA &  0.08 &                 0.12  &         \textbf{0.21} &                 0.10  &          0.12  &  \underline{\it 0.18} &                 0.11   &  0.05 \\
MU &  0.14 &                 0.15  &  \underline{\it 0.21} &                 0.18  &          0.13  &                 0.17  &                 0.17   &  \textbf{0.23} \\
PT & \underline{\it -0.09} & \textbf{-0.08} &        \textbf{-0.08} &                -0.21  &         -0.20  &                -0.15  &                -0.19   & \underline{\it -0.09} \\
SB &  0.26 &                 0.29  &  \underline{\it 0.44} &                 0.26  &          0.30  &                 0.32  &                 0.31   &  \textbf{0.47}\\
SH &  \underline{\it 0.09} &                 \underline{\it 0.09}  &  \underline{\it 0.09} &                 \underline{\it 0.09}  &          \underline{\it 0.09}  &                 0.08  &                 \underline{\it 0.09}   &  \textbf{0.10} \\
WI &  \underline{\it 0.29} &                 0.25  &  \textbf{ 0.42} &                 0.25  &          0.27  &                 0.24  &                 \underline{\it 0.29}   &  0.28\\
ZO &  0.38 &                 0.37  &                 0.41  &                 0.33  &          0.36  &                 0.38  &  \underline{\it 0.42}  &  \textbf{0.46} \\
\bottomrule
\end{tabular}
\end{table}

\subsection{Results\label{sec:direct}}

In this subsection, we directly verify the accuracy of the NECA algorithm and compare its performance against benchmark models. 

The internal cluster validity indices are used to quantitatively evaluate the effectiveness of the learned numerical representations, by measuring the compactness of intra-cluster and separation of inter-clusters. If two data objects belong to the same class, their learned representations should be similar, otherwise, the learned representations are expected to be dissimilar. 

Calinski-Harabasz index (CH) \cite{Ref153} and Silhouette index (S) \cite{Ref10}, are used to evaluate the representation learning performance in this paper. 
Specifically, the CH index is calculated as follows:
\begin{equation}\label{eqn:3-28}
CH(\pi) = \frac{ \frac{1}{T-1} \sum_{i=1}^{T} |C_i| d^2(c_i,c) }
               { \frac{1}{n-T} \sum_{j=1}^{T}\sum_{g=1}^{|C_j|} d^2(x_{c_j}^g,c_j) },
\end{equation}
where $n$ is the number of data objects, $T$ is the number of classes, $|C_i|$ is the number of data objects in the $i$-th class, $c_i$ is the centroid of the $i$-th class, $c$ is the global center of the entire dataset, $x_{c_j}^g$ is the $g$-th data object in the $j$-th class. The numerator of \autoref{eqn:3-28} is used to measure the degree of separation between classes, and the denominator is used to measure the closeness within classes. Greater CH value indicates better performance.

Additionally, the S index is calculated as follows:
\begin{equation}\label{eqn:3-29}
S(\pi) = \frac{1}{T} \sum_{i=1}^{T} \left( \frac{1}{|C_i|} \sum_{g=1}^{C_i} \frac{b(x_{c_j}^g)-a(x_{c_j}^g)}{max \{b(x_{c_j}^g), a(x_{c_j}^g) \} } \right),
\end{equation}
where $a(x_{C_i}^g)$ and $b(x_{C_i}^g)$ are used to respectively reflect the compactness of intra-cluster and separation of inter-clusters of the data object $x_{C_i}^g$ \cite{Ref10}. The larger the S value, the more accurate the representation learning, which means the learned numerical representations of two data objects within the same class are adjacent and those belong to different classes are nonadjacent.

The CH and S indices of different representation learning models on the 11 datasets are reported in \autoref{tab:3-3} and \autoref{tab:3-4}, respectively. 
Each comparative models are conducted five times on their optimal parameter configurations. And we report the best result of each model on each dataset, where
the best performance is shown in bold, and the next best is shown in italics and underlined.

\autoref{tab:3-3} indicates that the deep representations learned by NECA show the superior performance on the CH index to benchmark models on six datasets, and the second best on one dataset. This indicates that, when evaluated with the CH index, our proposed NECA algorithm can effectively capture the potential useful information hidden in CADs, and express it in the numerical representations. Additionally, CDE had the best performance on the four datasets, which means tha CDE is a fairly competitive model, only inferior to our proposed NECA.

\autoref{tab:3-4} shows the evaluation results on the S index. It should be noted that both CDE and our proposed NECA perform well, they produce the optimal learning results on five datasets, respectively, which means from the perspective of S, the capability of NECA and CDE to learn the numerical representation of categorical data is close, and our NECA can effectively handle the situations that CDE cannot deal with. Additionally, the performance of other comparative baselines is relatively inferior.

In summary, compared with the existing categorical data representation learning models and network embedding models not specially used for categorical data representation learning, the NECA algorithm proposed in this paper can learn the most accurate deep numerical representations of CAD, which contain the useful information hidden in the original dataset, and are in line with the actual class distribution of the dataset.

\section{Related Work\label{sec:relate}}

The existing related studies can be divided into two subgroups: direct encoding methods and embedding-based methods.

Direct encoding methods for CAD refers to the numerical symbolization of categorical attribute values. 
There are three commonly used direct encoding methods: numerical encoding, frequency-based encoding, and one-hot encoding.
\textit{Numerical encoding} 
assigns a different number to each unique category \cite{HAN201283}. 
The exact numbering scale can be application dependent. For example, one can choose \{0, 1, 2\}, or \{10, 20, and 25\}, as long as a different number is used to represent a different attribute value. 
This method creates numerical representation for CAD in a subjective manner. It requires domain insights and experimentation to find the most suitable numbering scale.
\textit{Frequency-based encoding} assigns a numerical value to a categorical attribute value based on the inverse of its frequency in the dataset \cite{Ref35}. 
For example, suppose that there are 100 records of job candidates data, in which ''Engineering (Specialty)'' appeared 40 times, then the frequency-based encoding for ''Engineering (Specialty)'' is designated as $\log(100/40)$.
This type of encoding assigns smaller values for very common categories and larger values for rare categories, which is consistent with the information theory \cite{Ref9}. 
Frequency-based encoding can be used for both ordinal and nominal attributes, 
but since it only considers frequency, it does not well differentiate categories with similar frequencies or reflect the actual relationship between different categories. 
\textit{One-hot encoding} is the most commonly used method for encoding CAD. It converts a categorical attribute into a set of binary attributes, where each corresponds to one possible category \cite{Ref33}. 
For any given data object, the categorical attribute then becomes a vector where only one entry (which corresponds to the binary attribute that represents that category) is 1, and all other entries are 0. 
Although one-hot encoding is highly effective and flexible for automatic encoding, 
it ignores the relationship between the attribute values and causes sparse problem. 

The embedding-based methods refer to incorporating a categorical data object into a low-dimensional continuous space to obtain a numerical representation that is similar to numerical data. 
There are several methods proposed in recent literatures. 
CDE \cite{Ref40} is a superior embedding-based method for categorical data representation learning. It uses metric learning \cite{Ref41} to dig into the relationship between CAVs, and introduces an clustering algorithm in the representation learning process. The CDE method was further improved into 
CDE++ \cite{Ref42}, which uses mutual information and marginal entropy to obtain the relationship between CAVs. It proposes a hybrid clustering strategy to capture the values of different types of categorical attributes clusters, then an automatic encoder (autoEncoder) \cite{Ref43} is employed to obtain a dense low-dimensional vector representation. 
The UNTIE method \cite{Ref44} is a shallow model, using frequency to represent the relationship between different CAVs, and the co-occurrence relationship is reflected by conditional probabilities, then the K-means algorithm is used to mine the heterogeneity between these relationships and obtain the embedding-based representations of CAD. 
Existing works on embedding-based methods consider the relationship between the categorical attributes and CAVs, which has advantages compared with the direct encoding methods.
However, they have the following limitations. 
Firstly, the clustering process is introduced into the representation learning, and the learning result is affected by the selection of clustering algorithms and the clustering parameters. 
Secondly, the representation learning process requires the use of external information, such as data labels, which is not suitable for unlabeled data that is often faced in real-world. 

\section{Conclusion\label{sec:conclusion}}

In this paper, 
we propose a novel categorical data representation learning algorithm, NECA, a deep representation learning method for categorical data based on network embedding.
Built upon the foundations of network embedding and deep unsupervised representation learning, NECA firstly constructs a weighted heterogeneous network based on the inter-categorical attribute relationship and intra-categorical attribute relationship. Next, the deep numerical representation of each categorical attribute value is learned by leveraging the basic idea of network embedding. Finally, the numerical representation of categorical data object can be produced through the concatenation operation. In this way, NECA deeply embeds the intrinsic relationships among categorical attribute values, and can explicitly express categorical data objects with the learned numerical representations, which contain the potential useful information hidden in the original dataset. 
Extensive experiments on 11 UCI datasets demonstrate that NECA outperforms several state-of-the-art benchmark models.

\bibliographystyle{unsrt}
\bibliography{refs}

\end{document}